\def\year{2020}\relax
\documentclass[letterpaper]{article} 
\usepackage{aaai20}  
\usepackage{times}  
\usepackage{helvet} 
\usepackage{courier}  
\usepackage[hyphens]{url}  
\usepackage{graphicx} 
\usepackage{graphicx}  
\usepackage{booktabs}       

\title{Label Augmentation via Time-based Knowledge Distillation for Financial Anomaly Detection}
\author{\textbf{Hongda Shen\textsuperscript{\rm 1}, Eren Kursun\textsuperscript{\rm 2}}\\ 
\textsuperscript{\rm 1}University of Alabama in Huntsville\\ 
hs0017@alumni.uah.edu \\
\textsuperscript{\rm 2}Columbia University\\ 
ek2925@columbia.edu
}
\begin{document}

\maketitle

\begin{abstract}
Detecting anomalies has become increasingly critical to the financial service industry. Anomalous events are often indicative of illegal activities such as fraud, identity theft, network intrusion, account takeover, and money laundering. Financial anomaly detection use cases face serious challenges due to the dynamic nature of the underlying patterns especially in adversarial environments such as constantly changing fraud tactics. While retraining the models with the new patterns is absolutely essential; keeping up with the rapid changes introduces other challenges as it moves the model away from older patterns or continuously grows the size of the training data. The resulting data growth is hard to manage and it reduces the agility of the models' response to the latest attacks. Due to the data size limitations and the need to track the latest patterns, older time periods are often dropped in practice, which in turn, causes vulnerabilities. In this study, we propose a label augmentation approach to utilize the learning from older models to boost the latest. Experimental results show that the proposed approach provides a significant reduction in training time, while providing potential performance improvement.

\end{abstract}
\section{Introduction}
Machine learning approaches for anomaly detection have found a wide range of application areas in financial services such as cyber defense systems, fraud detection, compliance and anti-money laundering \cite{AnandakrishnanKS2017}. Among these, there is a sizable list of mission-critical applications each of which requires effective and timely detection of anomalous events in real-time. In applications such as payment fraud detection systems, where tens of millions of transactions per day are scored with millisecond range response time SLAs, the underlying modeling challenges become more prominent.

One of the grand challenges in such systems is the adversarial nature of the detection process. Unlike data-sets where the underlying patterns are naturally stable, fraud and anomaly detection use cases typically deal with constant and often rapid changes \cite{MarfaingG2018}. The pattern changes occur in both (i) \textit{normal events}, as in changes in normal transactions and customer behavior, as well as (ii) \textit{anomalous events}, as in perpetrators implementing new fraud tactics in response to recent prevention measures.  For instance, Account Takeover (ATO) fraud typically involves fraudsters gaining access to customers account and draining the funds across multiple channels. ATO fraud tactics are known to show rapid changes. In some cases, perpetrators move from one popular tactic to the next in a matter of days.

In such dynamic and adversarial environments, machine learning especially supervised learning algorithms face a dilemma. While the retraining of the models with the new patterns improves the performance for recent trends, it frequently degrades the performance for historical patterns that may repeat. Excluding historical patterns causes retention challenges. Yet, continuously extending the training data set with additional data causes data size and training time issues. 

In this paper, we propose a novel supervised learning approach that provides a balance between these two opposing forces. This technique, Label Augmentation via Time-based Knowledge Distillation (LATKD) aims to transfer knowledge from historical data to boost the model through data labeling. The proposed solution improves the training time for agile response in adversarial use cases, such as  fraud detection and account takeover, as well as providing robust performance by combining a wider range of patterns over time.

\section{Related Work}
The concept of Knowledge Distillation (KD) was explored by a number of researchers  \cite{BuciluaC2006,BaJ2014,HintonG2015,UrbanG2016,FurlanelloT2018}. Initially, the goal of KD was to produce a compact student model that retains the performance of a more complex teacher model that takes up more space and/or requires more computation to make predictions. \textit{Dark Knowledge} \cite{HintonG2015}, which includes a softmax distribution of the teacher model, was first proposed to guide the student model. Recently, the focus of this line of research has shifted from model compression to label augmentation which can be considered a form of regularizer using \textit{Dark Knowledge}. In \cite{FurlanelloT2018}, \textit{Born Again Network} (BAN), a chain of retraining models, parameterized identically to their teachers, was proposed. The final ensemble of all trained models can outperform their teacher network significantly on computer vision and NLP tasks. Additionally, \cite{FurlanelloT2018} investigated the importance of each term to quantify the contribution of dark knowledge to the success of KD. Following this direction of research, self distillation has emerged as a new technique to improve the classification performance of the teacher model rather than merely mitigating computational or deployment burden. Label refinery \cite{BagherinezhadH2018} iteratively updates the ground truth labels after cropping the entire image dataset and generates a set of informative, collective, and dynamic labels from which one can learn a more robust model. In another related study, \cite{RomeroA2014} aimed to compress models by approximating the mapping between hidden layers of the teacher and the student models, using linear projection layers to train relatively narrower students.

In this study, we propose a label augmentation approach that incorporates \textit{Dark Knowledge} from previously trained models, which have been trained with different time ranges to augment the labels of the latest dataset. This new knowledge enables the transfer of learning from historical patterns extracted by experienced \textit{experts}. With the assistance of their expertise, the new model sees performance improvement without having the historical data-sets in its training. This enables more effective detection of anomalous events, and streamlines model retraining and deployment.

\section{LATKD: Label Augmentation via Time-based Knowledge Distillation}
Consider the classical classification setting with a sequence of training datasets corresponding to $N$ different time frames consisting feature vectors: $X_t$ and labels $Y_t$ where $t=0,1,...N$. For traditional supervised learning algorithms, a model is trained on $\{X_{<t},Y_{<t}\}$ for each time frame. Naturally, the size of $\{X_{<t},Y_{<t}\}$ increases as time passes. LATKD leverages the outputs generated by previously trained models $M_{<t}$  prior to each time frame $t$ instead of including historical data in the training directly. These outputs are used to augment labels of the latest dataset and construct a regularizer to the conventional loss function. For time frame $t$, the training dataset will be $\{X_{t},Y_{t}\}$ only and the loss function to optimize in the training becomes:

\begin{equation}
\label{eq:loss}
Loss_{t} =  CE(Y_t, y_t)+ \sum_{i=K}^{t-1} KL(O_{i,t},y_t)
\end{equation}
where $O_{i,t}$ and $y_t$ represents model $M_i$ output on data $X_t$ and model output at the current time frame, respectively.  $CE$ and $KL$ are Cross-Entropy and Kullback–Leibler divergence. With this second term in the loss function Eq. \ref{eq:loss}, existing ground truth labels are augmented by the \textit{experienced experts}. 

As the number of models increases over time, the historical models, whose underlying training data patterns have changed provide increasingly less meaningful information on the recent anomaly patterns. Thus, including them in the training may not provide further performance gain for retraining and possibly deteriorate the performance. To reduce the negative impact of this distribution shift, we use parameter $K$ to determine which model to start with and truncate all the previous models prior to the current one. In this study we used an empirical approach to determine $K$.

\begin{figure}[h]
  \centering
  \includegraphics[scale=0.63]{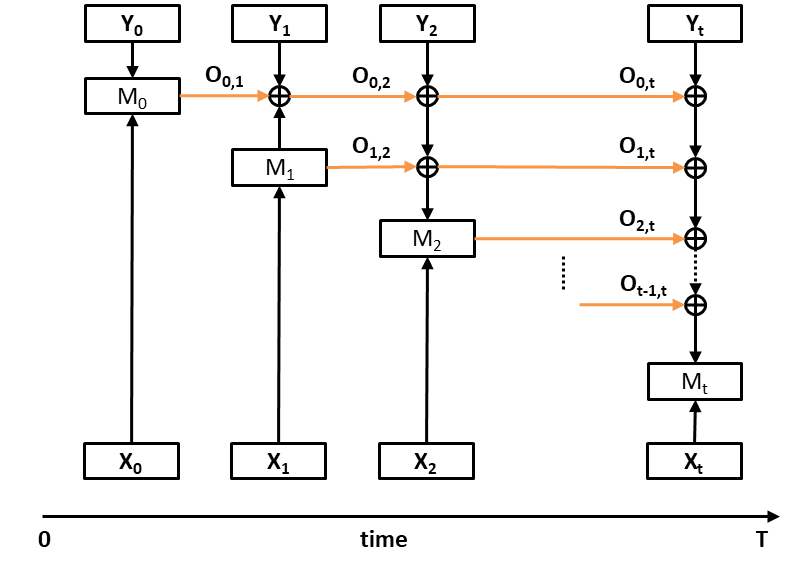}
  \vspace{-1em}
  \caption{Architecture of Label Augmentation via Time-based Knowledge Distillation (LATKD).}
\label{fig:diagram}
\vspace{0em}
\end{figure}

Fig. \ref{fig:diagram} illustrates the architecture of LATKD. For the first time frame $t=0$, a model $M_0$ is trained on dataset $\{X_{0},Y_{0}\}$. Then, for each of the following time frames (depending on the specific retraining schedule), a new identical model $M_t$ is trained from, $O_{i,t}$ the supervision of previous models $M_{<t}$ by using Eq. \ref{eq:loss}. Auxiliary labels (outputs) from the previous models are highlighted in orange in Fig. \ref{fig:diagram}.

\section{Experimental Analysis}
This section provides the experimental analysis for LATKD using an open-source anomaly detection dataset \cite{IEEE_CIS_Dataset} based on telecommunications industry card-not-present payment transactions. As in almost all the anomaly detection problems, negative class in this data set takes a very small portion of the total transactions. For the experimental analysis, we extracted 6 months of data with the labels included. The first day of this data set is assumed to be November 1st, 2017 \cite{IEEE_CIS_Dataset_Timeframe_Analysis}. The start date was used to facilitate data segmentation and does not impact the model performance. November 2017 - January 2018, was used as the training period while March - April 2018 was used as the testing period. Data, including labels from additional months, were gradually added in increments of 1 month into the training starting with November 2017 to focus on an adversarial fraud detection environment with monthly training. Table \ref{tab:exp_runs} shows further details for each experiment period.

\begin{table}[h]
  \vspace{-1em}
  \caption{Experimental periods details.}
  \label{tab:exp_runs}
  \centering
  \scalebox{0.7}{
  \begin{tabular}{cllr}
    \toprule
    Period \#  & Training Period   & Testing Period & Training \# Nonfraud / \# Fraud \\
    \midrule
    1 & Nov. & Mar. + Apr. & $130937$ / $3401$  \\
    2 & Nov. + Dec. & Mar. + Apr. & $219758$ / $7090$  \\
    3 & Nov. + Dec. + Jan. & Mar. + Apr.  & $315156$ / $11029$ \\
    \bottomrule
  \end{tabular}}
\end{table}

We assume a 30-day delay for data labeling to account for claim submission process and labeling. Therefore, February 2018 is considered as unlabeled; hence it was not used for training. Categorical features were encoded using \textit{one-hot encoding}. \textit{log10} transformation was used on continuous variables to limit their value ranges. Further details on feature preprocessing can be found in Table \ref{tab:dataset_preprocessing} in the Appendix. Area Under Precision-Recall Curve (AUPRC) was selected to compare classification performance as the primary metric. AUPRC has been shown as a stronger metric for performance and class separation than Area Under Receiver Operating Curve (AUROC) in highly imbalanced binary classification problems \cite{JesseD2006,SaitoT2015}.

In this section, we demonstrate the effectiveness of the proposed approach and conduct a comparison between the baseline of commonly used machine learning approaches and the corresponding LATKD versions: (i) \textit{MLP}: A Multi-layer Perceptron based architecture has been trained on labeled data to serve as the baseline. Implementation details of the MLP has been provided in Table \ref{tab:mlp} in Appendix (ii) \textit{XG}: Xgboost algorithm \cite{ChenT2016} is a variant of Gradient Boosting Trees which has been widely used to model tabular data (from Kaggle competitions to industrial applications) due to its high efficiency and performance. Specific set of hyperparameters for this study were determined using grid search and provided in Table \ref{tab:xg} in Appendix (iii) \textit{MLP-XG}: An ensemble of baseline Xgboost and MLP via averaging outputs of both models (iv) \textit{MLP-XG-LATKD}: Label Augmented MLP-XG using historical ensemble models.


\begin{figure}[h]
  \centering
  \includegraphics[scale=0.48]{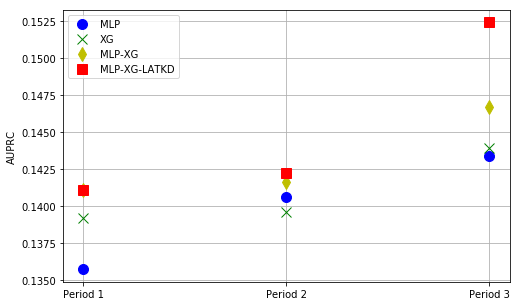}
    \vspace{-1em}
  \caption{AUPRC for MLP, XG, MLP-XG and MLP-XG-LATKD.
}
\vspace{0em}
\label{fig:aucpr}
\end{figure}

We use a supervised binary classification approach, where each algorithm was run 10 times for each training time period. AUPRC value for each run is collected and the average of all the 10 collected values is recorded as the final performance measure.

\begin{table}[h]
  \vspace{-1em}
  \caption{Relative AUPRC difference of experimented methods against baseline MLP.}
  \label{tab:performance}
  \centering
  \scalebox{0.7}{
  \begin{tabular}{crrr}
    \toprule
    Period \# & XG & MLP-XG & MLP-XG-LATKD \\
    \midrule
    1 & 2.28\% & 3.70\% & \textbf{3.70\%}\\
    2 & -0.71\% & 0.70\% & \textbf{1.14\%}\\
    3 & 0.35\% & 2.31\% & \textbf{6.31\%}\\
    \bottomrule
  \end{tabular}}
\end{table}

Fig. \ref{fig:aucpr} shows the AUPRC for the aforementioned methods over three experiment periods.  AUPRC improvement over baseline MLP is shown in Table \ref{tab:performance}. XG outperformed MLP for Period 1 and Period 3 while MLP performed better in Period 2. Furthermore, the ensemble of MLP and XG, MLP-XG, outperformed both models by 2.23\% on average and up to 3.7\% on AUPRC improvement against the baseline MLP. LATKD augmented MLP-XG produced the best performance for all three models. Particularly, in Period 3, by having two previous models to augment the labels, LATKD presented significantly better performance over baselines. Similar performance improvement was observed by applying LATKD on MLP and XG separately. 

\begin{figure}[h]
  \centering
  \includegraphics[scale=0.48]{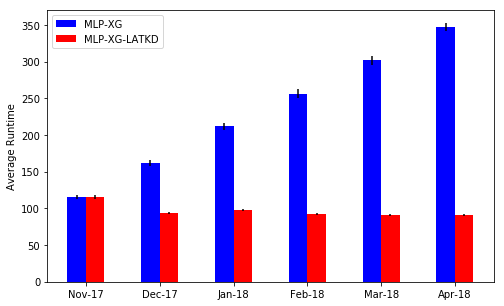}
    \vspace{-1em}
  \caption{Average runtime comparison between MLP and MLP-LATKD.
}
\vspace{0em}
\label{fig:runtime}
\end{figure}

From the performance comparison, MLP-XG ensemble and MLP-XG-LATKD were identified as the highest performance approaches. Fig.\ref{fig:runtime} shows the average runtime in seconds for MLP-XG and MLP-XG-LATKD over 10 repeated runs from November 2017 to April 2018. A machine with Intel (R) Core (TM) i7-6700HQ CPU at 2.6GHz, 16GB RAM and NVIDIA GTX 960M GPU was used for the runtime comparison. MLP was trained with cumulative time periods of data (similar to Table \ref{tab:exp_runs}) while the training period of LATKD only included the month itself without any historical data. An extended version of the time range up to Apr-18 was used to better illustrate the LATKD runtime advantage over the time. 

Since LATKD enables the transfer of learning from historical data, only limited recent training period was used to train each model. As a result, the average runtime only depends on the size of the latest data set. On the other hand, traditional supervised learning techniques including both MLP and XG require all the available data in their training, which leads to super-linear increases in the training time. Blue and red bars stand for MLP-XG and MLP-XG-LATKD average training times in Fig. \ref{fig:runtime}. Both methods take the same time to run at the beginning. Gradually, with more data added in MLP training, its runtime increases while runtime of MLP-XG-LATKD remains approximately the same. MLP-XG-LATKD provides lower runtimes consistently over Dec-17 through Apr-18. Over the 6 months experimentation period, the average runtime was reduced by 58.5\% with up to 3.8x improvement in Apr-18. It is important to note that the training runtime advantage of LATKD shown in this experiment translates to significantly higher numbers in real-life implementations with larger data sets, further yielding reduced runtime, and resources. This, in turn, yields improved training time, computational cost and agility of responses in adversarial environments. LATKD provides the opportunity to boost the performance of the individual models as well as the ensembled models.

\section{Conclusions}
In this study, we propose, LATKD, a label augmentation algorithm for financial anomaly detection applications. LATKD provides a way to boost the model performance by incorporating a wider range of patterns including older and newer patterns without unmanageably increasing the data set size, while maintaining a robust performance. In adversarial and time-critical use cases such as cyber defense, account takeover fraud this provides significantly higher agility and a more effective response to attacks.

\bibliography{reference}
\bibliographystyle{aaai}

\section{Appendix}
\begin{table}[h]
  \caption{Dataset Preprocessing Details}
  \label{tab:dataset_preprocessing}
  \centering
  \scalebox{0.65}{
  \begin{tabular}{lllll}
    \toprule
    Raw feature  & Type   & Encoding & Null value & Notes \\
    \midrule
    TransactionAmt & Continuous & $log10()$  & - &   -   \\
    dist1 & Continuous & $log10()$ & $-0.001$ &  -    \\
    dist2 & Continuous & $log10()$  & $-0.001$ &  - \\
    ProductCD & Categorical & One hot  & - & - \\
    card4 & Categorical & One hot  & NA  & -\\
    card6 & Categorical & One hot  & NA  & -\\
    M1-M9 & Categorical & One hot  & NA  & -\\
    device\_name & Categorical & One hot  & NA  & ``Others" if frequency $<$ 200\\
    OS & Categorical & One hot  & NA  & -\\
    Browser & Categorical & One hot  & NA  & ``Others" if frequency $<$ 200\\
    DeviceType & Categorical & One hot  & NA  & -\\
    \bottomrule
  \end{tabular}}
\end{table}

\begin{table}[h]
  \caption{Multi-layer Perceptron Architecture}
  \label{tab:mlp}
  \centering
  \scalebox{0.77}{
  \begin{tabular}{llll}
    \toprule
    Layer  & \# Neurons  & Activation function & Parameter \\
    \midrule
    Dense & 400 & RELU & -   \\
    BatchNormalization & - & - & -   \\
    Dropout & - & -  & keep\_prob = 0.5 \\
    Dense & 400 & RELU  & -  \\
    Dropout & - & -  & keep\_prob = 0.5\\
    Dense (Output) & 2 & Softmax & -\\
    \midrule
    learning rate & - & - & 0.01\\
    Batch size & - & - & 512\\
    \bottomrule
  \end{tabular}}
\end{table}

\begin{table}[h]
  \caption{Xgboost Hyperparameters}
  \label{tab:xg}
  \centering
  \scalebox{0.8}{
  \begin{tabular}{ll}
    \toprule
    Name  & Value  \\
    \midrule
    colsample\_bytree & 0.8   \\
    gamma & 0.9   \\
    max\_depth & 3 \\
    min\_child\_weight & 2.89  \\
    reg\_alpha & 3\\
    reg\_lambda & 40\\    
    subsample & 0.94\\
    \midrule
    learning\_rate & 0.1\\
    n\_estimators & 200\\
    \bottomrule
  \end{tabular}}
\end{table}

\end{document}